\documentclass[letterpaper, 10 pt, conference]{ieeeconf}  

\IEEEoverridecommandlockouts                              

\usepackage{graphicx}
\usepackage{times}
\usepackage{epsfig}
\usepackage{graphicx}
\usepackage{amsmath}
\usepackage{amssymb}
\usepackage[breaklinks=true,bookmarks=false]{hyperref}

\title{\LARGE \bf
Driving Experience Transfer Method for End-to-End Control of Self-Driving Cars
}

\author{Dooseop Choi, Taeg-Hyun An, Kyounghwan Ahn, and Jeongdan Choi
\thanks{This work was supported by Institute for Information and communications Technology Promotion (IITP) grant funded by the Korea government (MSIP) [2017-0-00068, A Development of Driving Decision Engine for Autonomous Driving(4th) using Driving Experience Information]}
\thanks{The authors are with Electronics and Telecommunications Research Institute, Daejeon, Republic of Korea.
{\tt\small $\{$d1024.choi, tekkeni, mobileguru, jdchoi$\}$@etri.re.kr }%
}
}

\begin{document}

\maketitle
\thispagestyle{empty}
\pagestyle{empty}

\begin{abstract}
In this paper, we present a transfer learning method for the end-to-end control of self-driving cars, which enables a convolutional neural network (CNN) trained on a source domain to be utilized for the same task in a different target domain. A conventional CNN for the end-to-end control is designed to map a single front-facing camera image to a steering command. To enable the transfer learning, we let the CNN produce not only a steering command but also a lane departure level (LDL) by adding a new task module, which takes the output of the last convolutional layer as input. The CNN trained on the source domain, called $\textit{source network}$, is then utilized to train another task module called $\textit{target network}$, which also takes the output of the last convolutional layer of the source network and is trained to produce a steering command for the target domain. The steering commands from the source and target network are finally merged according to the LDL and the merged command is utilized for controlling a car in the target domain. To demonstrate the effectiveness of the proposed method, we utilized two simulators, TORCS and GTAV, for the source and the target domains, respectively. Experimental results show that the proposed method outperforms other baseline methods in terms of stable and safe control of cars.
\end{abstract}

\section{INTRODUCTION}

Autonomous driving has long been a great interest in both academia and industry. A variety of technologies such as object detection and tracking, localization, and path planning must be simultaneously considered in order to enable a car to drive itself. In the past few decades, little progress had been made in autonomous driving due to the lack of the performance breakthrough of the conventional technologies. Significant progress in autonomous driving has been made in the past few years owning to the success of deep convolutional neural networks (CNNs) in computer vision tasks. The success of deep CNNs is attributed to the availability of millions of training samples as well as the advances in network architecture and training techniques. 

The success not only advances each technology required for autonomous driving, but also motivates researchers to propose new paradigms \cite{Chen}\cite{Bojarski}\cite{Xu}\cite{Yu}, and controlling self-driving cars in an end-to-end fashion is one of the paradigms. Motivated by \cite{Pomerleau}, Bojarski et al. \cite{Bojarski} proposed training a deep CNN to map a single front-facing camera image to a steering command so that there is no need to carefully re-design and tune each technology required for autonomous driving whenever the self-driving car faces a new driving environment. 

For the successful driving, however, the data required for the training of the deep CNN needs to be collected carefully as illustrated in Figure 1. First, the steering commands in the training dataset should be distributed as uniformly as possible over all possible steering commands so that the trained CNN can drive on straight roads as well as sharply curved roads. Second, the dataset should include the experience of returning to the lane center when a car is out of its lane for the safe driving. However, collecting such data in real world can be expensive, time-consuming and dangerous. (footnote: write why it is expensive, time-consuming, and dangerous.)
\begin{figure}[t]
\centerline {\includegraphics[width=5.0cm]{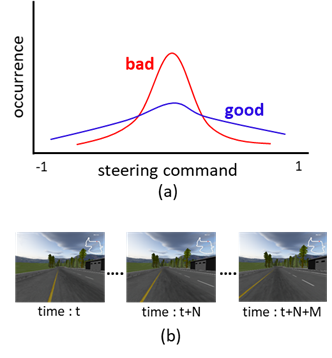}}
\caption{ Examples of (a) the distribution of the steering commands in a training dataset and (b) the experience of returning to the lane center.}
\end{figure}

\begin{table*}[t]
\centering
\caption{The structure of the proposed network. Relu and LRN are abbreviations for rectified linear unit and local response normalization, respectively.}
\label{table1}
  \begin{tabular}{| c c c c c c c c c c | }
    \hline
    Layer    		& Conv1 		& Conv2 		& Conv3 		& Conv4 		& Conv5 		& Fc6 & Fc7 & Fc8 & Fc9 \\ \hline \hline
    Kernel Size   	& (5$\times$5)  &(5$\times$5)   &(5$\times$5)  	&(3$\times$3)	&(3$\times$3)	& -   & -   & -   & -   \\ \hline
    Stride   		& (2,2)  		&(2,2)   		&(2,2) 		 	&(1,1)			&(1,1)	& -   & -   & -   & -   \\ \hline
    Activation   	& Relu  		& Relu   		& Relu 		 	& Relu			& Relu			& Relu   & Relu   & Relu   & Tanh   \\ \hline    
	Additional   	& LRN  			& LRN   		& LRN 		 	& -				& -				& -   	& -   & -   & -   \\ \hline        
  \end{tabular}
\end{table*}
\begin{figure*}[t]
\centerline {\includegraphics[width=12.0cm]{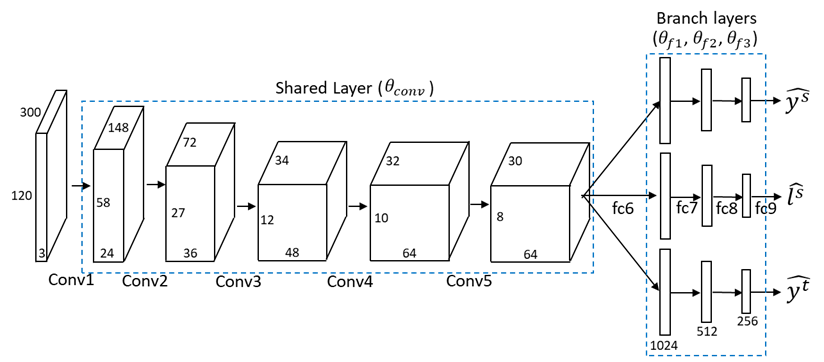}}
\caption{ The shared layer ($\theta_{conv}$) and the three task modules ($\theta_{f1}$, $\theta_{f2}$, $\theta_{f3}$).}
\end{figure*}

One alternative way to solve this problem is to collect data from simulators and use it for real-world tasks. Let us denote a simulator and real-world as the source domain and the target domain, respectively. Many researches have been proposed to utilize the dataset from the source domain (source dataset) for the tasks in the target domain, and most of them are focused on object classification, detection, and segmentation \cite{Daume1,Daume2,Aytar,Fernando,Chopra,Hoffman,Tzeng,Fernando2,Bousmalis}. Most of these can be categorized into four major paradigms: 1) train a network for the task on the target domain (called $\textit{target network}$) with the union of the source and target datasets, 2) utilize the outputs or the low and mid-level features of the network, which is trained with the source dataset (called $\textit{source network}$), as a feature for the target network, 3) train a target network with the target dataset under the restriction that the parameters of the target network should be as close as possible to those of the source network, 4) train a target network such that the network well recognizes common information between the source and the target data.\\

The proposed transfer learning method may fall into the second category. The mid-level features (the output of the last convolutional layer) as well as the output steering command of the source network are used to not only train a target network but also control a car in the target domain. We choose the second paradigm since it is found from our experiments that a well trained source network recognizes common information (lane lines which are crucial factors for steering) between images from the source and target domain very well. We will show how the source network recognizes images from the source and target domain in the next section. The structure of the proposed network as well as its training method are based on our observation and analysis.

Finally, it is worthwhile to note that Pan et al. \cite{Pan} also tried to utilize the source dataset for the target domain task focusing on the end-to-end control of self-driving cars. They trained a deep network that translates an image from the source dataset to the corresponding image in the target domain. Their approach, however, has a critical limitation that only three actions (go straight, turn left, turn right) can be estimated from the transfer learning. In contrast, our network directly produces a steering command from an input image in the target domain.

The rest of this paper is organized as follow. In section II, we first define the goal we want to achieve and then propose a CNN and its training method for driving experience sharing. Several CNNs trained with different methods are compared in section III to evaluate the proposed method and some conclusions are provided in section IV.
\begin{figure*}[t]
\centerline {\includegraphics[width=12.0cm]{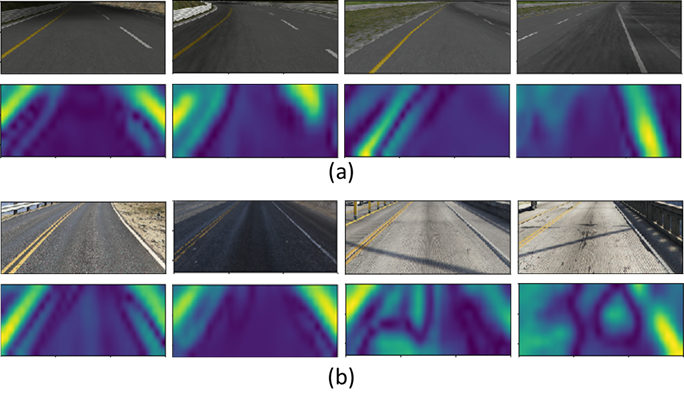}}
\caption{Road images (upper row) and corresponding activation maps (bottom row). The road images in (a) and (b) are collected from TORCS and GTAV, respectively. }
\end{figure*}

\section{Proposed transfer learning method}
\subsection{Problem Statement}
Let $\mathbf{S} = \{(\mathbf{x}_{s}(i), \mathbf{y}_{s}(i), \mathbf{l}_{s}(i)) \}^{N_{s}}_{i=1}$ denote a dataset of $N_{s}$ samples from the source domain where $\mathbf{x}_{s}(i)$ is a front-facing camera image, $\mathbf{y}_{s}(i)$ ($\in \mathbb{R}^{1}$) and $\mathbf{l}_{s}(i)$ ($\in \mathbb{R}^{1}$) are the corresponding steering command and normalized lateral distance between the car center and lane center line, respectively. We call $\mathbf{l}_{s}(i)$ lane departure level (LDL). Also let $\mathbf{T} = \{(\mathbf{x}_{t}(i), \mathbf{y}_{t}(i))\}^{N_{t}}_{i=1}$ denote a dataset of $N_{t}$ samples from the target domain. In this paper, it is assumed that the source dataset satisfies the two conditions mentioned in section I so a CNN trained with the source dataset well drives on roads in the source domain. On the other hand, the target dataset doesn't satisfy at least one of the two conditions. Our goal is then to train a CNN, which can drive a car in the target domain in a stable and safe manner, with the source dataset $\mathbf{S}$ and the target dataset $\mathbf{T}$. 

In this paper, for the source and the target domains, we utilized two simulators, TORCS \cite{Wymann} and GTAV. TORCS was used for the source domain because all the information about cars and driving environments can be obtained while driving. On the other hand, GTAV provides realistic driving environments and various types of loads so we used it for the target domain. 

Finally note that, in the remainder of this paper, we assume that a car drives on roads with lane lines. Also the sample index $i$ is omitted for the sake of simplicity.

\subsection{Network}
The proposed network is the modified version of the one proposed in \cite{Bojarski}. The main difference is that the proposed has two additional modules right after the last convolutional layers. Figure 2 and Table 1 show the structure of the proposed network in detail. The input to the network is an RGB image of size 300$\times$120 pixels, which is normalized by its pixel mean and variance. The outputs of the network are two steering wheel commands ($\hat{\mathbf{y}}^{s}$, $\hat{\mathbf{y}}^{t}$) and one lane departure level (LDL, $\hat{\mathbf{l}}^{s}$), which are real values and ranging from -1.0 to 1.0. For examples, $\hat{\mathbf{y}}=1.0$ and $\hat{\mathbf{l}}=1.0$ respectively mean that the steering wheel is turned all the way to the right and that the car is located midway between the current and right lanes.

The five convolutional layers (parameterized by $\theta_{conv}$) of the network are shared by three task modules. The fully connected layers of the first, second, and third task modules are parameterized by $\theta_{f1}$, $\theta_{f2}$, and $\theta_{f3}$, respectively. The convolutional layers as well as the fully connected layers of the first and second modules are trained only with $\mathbf{S}$. The fully connected layers of the third module is then trained with $\mathbf{T}$ given the optimized parameters ($\theta_{conv}$, $\theta_{f1}$, $\theta_{f2}$). This sequential approach is based on our observation that the network trained with $\mathbf{S}$ well recognizes common information, which is required for steering such as lane lines, between the source and the target images. Figure 3 shows examples. The upper rows of Figure 3-(a) and (b) show the source and the target images, respectively, while the bottom rows are the corresponding activation maps obtained from the output of $\textit{Conv5}$ layer of the source network. The brightest yellow on the map indicates the highest activation while the darkest blue indicates the lowest activation. We obtained the activation maps by averaging the output of the $\textit{Conv5}$ layer along the channel axis. The weights for the average were determined by the optimized parameters of the last fully connected layer, which are used for the final steering command prediction, as presented in \cite{Zhou}. As a result, it can be said that the prediction of the steering command is mainly influenced by the input image areas with high activations.

It is seen in the figure that high activations occurred mainly in the vicinity of the lane lines when the inputs to the network were the source images or the target images, which are similar to the source images. When the target images, which the source network is unfamiliar with, were input, high activations occurred not only in the vicinity of the lane lines but also around the lane center. (See the third and fourth columns from the left of the figure 3-(b).) However, we found from our experiment that, even with the target images, which the source network is unfamiliar with, the highest activation occurred in the vicinity of the lane lines, and such a tendency became noticeable as the car approaches the lane lines.

In the next section, we describe the proposed training method in detail.
\begin{figure*}[t]
\centerline {\includegraphics[width=14.0cm]{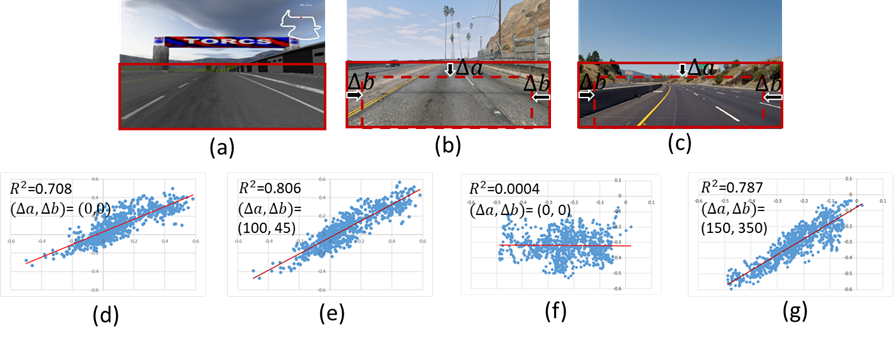}}
\caption{Front-facing camera images of (a) TORCS (640$\times$480), (b) GTAV (1920$\times$1080), (c) real-world (1280$\times$720). (d)-(g) Ground-truth LDL versus $\hat{\mathbf{l}}_{t}^{s}$ obtained from GTAV images ((d) and (e)) and real-world images ((f) and (g)). The red lines are the fitting results.}
\end{figure*}
\begin{figure}[t]
\centerline {\includegraphics[width=8.0cm]{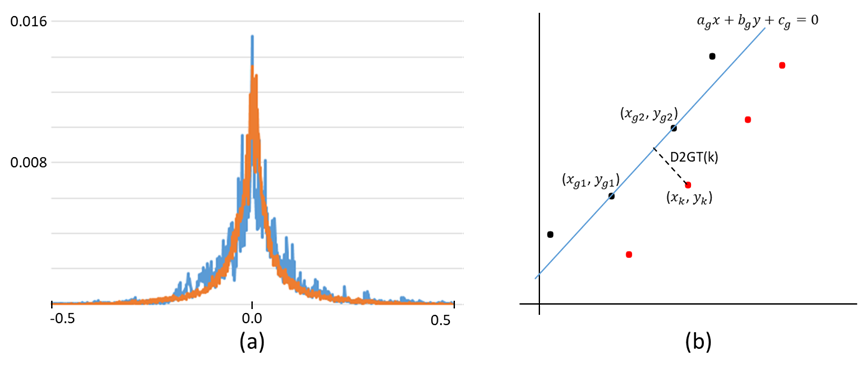}}
\caption{(a) the normalized histogram of the steering commands in the source dataset (blue) and the target dataset (orange), (b) example of calculating D2GT at time index $k$.}
\end{figure}
\begin{figure}[t]
\centerline {\includegraphics[width=8.0cm]{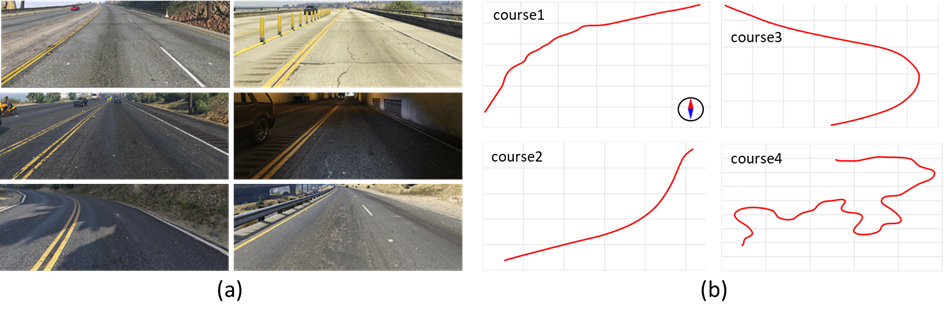}}
\caption{Example road images (a) and bird's-eye views (b) of the four courses in the target domain (GTAV).}
\end{figure}

\subsection{Training}
The convolutional layers ($\theta_{conv}$) as well as the first two task modules ($\theta_{f1}$, $\theta_{f2}$) are first trained with $\mathbf{S}$. The corresponding parameters can be optimized simultaneously by using the following loss function.
\begin{multline} 
\mathcal{L}_{1}(\mathbf{x}_{s}, \mathbf{y}_{s}, \mathbf{l}_{s}; \theta_{conv}, \theta_{f1}, \theta_{f2}) = 
||\mathbf{y}_{s} - f(\mathbf{x}_{s};\theta_{conv}, \theta_{f1})||^{2} + \\ \lambda_{1}||\mathbf{l}_{s} - f(\mathbf{x}_{s};\theta_{conv}, \theta_{f2})||^{2},
\end{multline} 
where $f(\mathbf{x};\theta)$ is a function parameterized by $\theta$, which maps an image $\mathbf{x}$ to a prediction value. $\lambda_{1}$ is a constant and we set $\lambda_{1}=1$ during our experiments.

Next, with the optimized parameters ($\hat{\theta}_{conv}$, $\hat{\theta}_{f2}$), the images in $\mathbf{T}$ are pre-processed by using  $\hat{\mathbf{l}}_{t}^{s}=f(\mathbf{x}_{t};\hat{\theta}_{conv}, \hat{\theta}_{f2})$ (the output LDL of the source network when a target image is input) in order to make the source network better understand the images. Let us describe in detail. For our experiments in section III, we obtained $\mathbf{x}_{s}$ by first cropping the bottom half of the images, which are rendered by TORCS and of size 640x480 pixels, and then resizing the cropped image to the size of 300x120 pixels as seen in Figure 4-(a). However, the images rendered by GTAV (Figure 4-(b)) or recorded by a video camera (Figure 4-(c)) may have different resolutions and camera field-of-view characteristics. Therefore, if we just use the bottom half of a target image for the input to the source network, the network may have difficulty understanding the input image. So we propose reducing the size of the cropping area progressively until $\hat{\mathbf{l}}_{t}^{s}$ shows the strongest positive correlation with the ground-truth LDL of the target images. Figure 4-(d) and (e) plot the ground-truth LDL versus $\hat{\mathbf{l}}_{t}^{s}$ over 760 examples from GTAV. We obtain the ground-truth LDL values manually with the aid of a traditional lane extraction algorithm, which utilizes edge, color, texture information of images. We can see in the figure that the strongest correlation does not occur at $\Delta a = 0$ and $\Delta b = 0$. We also conducted additional experiments using images from real-world, which are of size 1270$\times$720 pixels and were calibrated by camera intrinsic parameters. Figure 4-(f) and (g) show the results. It is seen in the figures that $\Delta a = 130$ and $\Delta b = 350$ results in stronger positive correlation than $\Delta a = 0$ and $\Delta b = 0$. Throughout this paper, whenever experiments need the target images as input to a network, they were pre-processed as described herein.

Finally, the third task module is trained with the target images and their corresponding steering commands under the involvement of the source network. Given the optimized parameters ($\hat{\theta}_{conv}$, $\hat{\theta}_{f1}$, $\hat{\theta}_{f2}$), $\theta_{f3}$ is optimized by minimizing 
\begin{multline} 
\mathcal{L}_{2}(\mathbf{x}_{t}, \mathbf{y}_{t}, \hat{\theta}_{conv}, \hat{\theta}_{f1}, \hat{\theta}_{f2}; \theta_{f3}) = \\ 
(1 - \varphi(|\hat{\mathbf{l}}_{t}^{s}|)) \times ||\mathbf{y}_{t} - f(\mathbf{x}_{t}; \hat{\theta}_{conv}, \theta_{f3})||^{2} \\ +  \varphi(|\hat{\mathbf{l}}_{t}^{s}|) \times  ||\hat{\mathbf{y}}_{t}^{s} - f(\mathbf{x}_{t}; \hat{\theta}_{conv}, \theta_{f3})||^{2},
\end{multline} 
where $\hat{\mathbf{y}}_{t}^{s} = f(\mathbf{x}_{t}; \hat{\theta}_{conv}, \hat{\theta}_{f1})$ and $\hat{\mathbf{l}}_{t}^{s} = f(\mathbf{x}_{t}; \hat{\theta}_{conv}, \hat{\theta}_{f2})$ are the steering command and the LDL value estimated by the source network when $\mathbf{x}_{t}$ is input, respectively. $\varphi(x)$ is an increasing function of $x$ and defined as $\varphi(x)=1/(1+\exp
(a\times(x-b)))$ in this paper. It is seen in (2) that $\theta_{f3}$ is optimized to produce a steering command close to $\hat{\mathbf{y}}_{t}^{s}$ when a car in the target domain turns out to be far from the lane center. Otherwise, it is optimized to produce one close to $\mathbf{y}_{t}$. The reason behind this design is that the steering command of the source network is more reliable as the car approaches the lane marking of its lane. (Recall our observation that the source network's attention to the line marking increases as the car approaches the marking.) As a result, through the loss function in (2), the target network ($\theta_{f3}$) can learn how to return to the lane center when the car is out of its lane.

\subsection{Driving}
The loss function in (2) may play the same role as $||\mathbf{y}_{t} - f(\mathbf{x}_{t}; \hat{\theta}_{conv}, \theta_{f3})||^{2}$ if the car was always on the center of the lane during the target data collection. Indeed, we drove a car as close as possible to the lane center when collecting the data. Nevertheless, the cases of deviating from the lane center occurred frequently while driving on curvy roads. Our experiments in section III showed that the car controlled by $\hat{\mathbf{y}}^{t}=f(\mathbf{x}; \hat{\theta}_{conv}, \hat{\theta}_{f3})$ drives well on straight and gently curved roads. However, when it drives on sharply curved roads, it often leaves its lane and fails to return. This is because most of the roads in the target domain are straight or gently curvy so the cases of deviating from the lane center take up a small part of the target dataset.

To deal with the situation where the car drives on sharply curved roads, we propose controlling the car with the following steering command.
\begin{equation} 
\hat{\mathbf{y}}^{share} = (1 - \varphi(|\hat{\mathbf{l}}^{s}|)) \times \hat{\mathbf{y}}^{t} +  \varphi(|\hat{\mathbf{l}}^{s}|) \times  \hat{\mathbf{y}}^{s},
\end{equation}
where $\hat{\mathbf{y}}^{s}=f(\mathbf{x}; \hat{\theta}_{conv}, \hat{\theta}_{f1})$ and $\hat{\mathbf{l}}^{s}=f(\mathbf{x}; \hat{\theta}_{conv}, \hat{\theta}_{f2})$. We let the parameters of $\varphi()$ in (2) and (3) be ($a=-12.0, b=-4.0$) and ($a=-8.0, b=-5.5$), respectively, for our experiments in section III. The difference between the two settings is that, with the latter setting, the source network is less involved in driving. 
\begin{figure*}[t]
\centerline {\includegraphics[width=17.0cm]{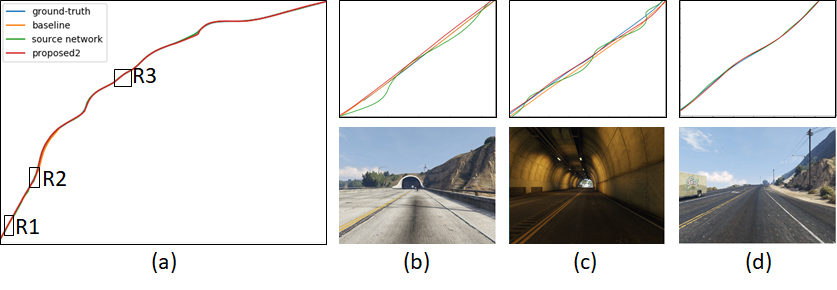}}
\caption{ Recorded car trajectories on $\textit{course 1}$. (a) whole trajectories. (b) trajectories on region $R1$ and corresponding road image. (c) trajectories on region $R2$ and corresponding road image. (d) trajectories on region $R3$ and corresponding road image.}
\end{figure*}
\begin{figure*}[t]
\centerline {\includegraphics[width=17.0cm]{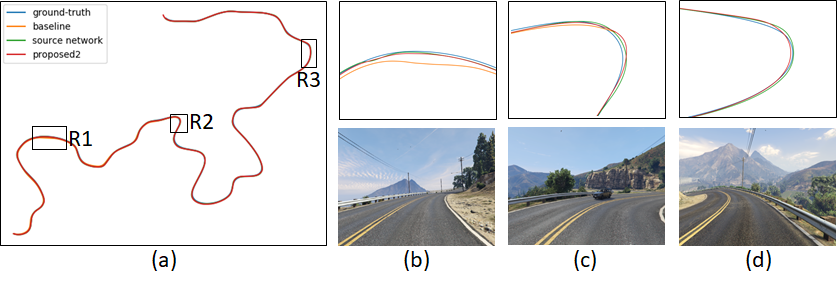}}
\caption{ Recorded car trajectories on $\textit{course 4}$. (a) whole trajectories. (b) trajectories on region $R1$ and corresponding road image. (c) trajectories on region $R2$ and corresponding road image. (d) trajectories on region $R3$ and corresponding road image.}
\end{figure*}

\section{Evaluation}

\subsection{Environments}
To construct $\mathbf{S}$ from TORCS, we made a racing robot that drives a car along the pre-defined paths on 10 tracks. Each track has 2 or 3 lanes and has a length of at least 2.5km. We collected 123,218 examples at a 10Hz sampling rate while the robot was driving at 40km/h. Finally, in order to make $\mathbf{S}$ satisfy the first condition mentioned in section I, 74,560 out of 123,218 examples were used for $\mathbf{S}$. Note that we carefully designed the pre-defined paths in order to make the source dataset satisfy the second condition mentioned in section I. For $\mathbf{T}$, we collected 122,516 examples at a 10Hz sampling rate while we were manually driving a car on the pre-defined courses of GTAV using Logitech G27 racing wheel. We used 68,333 out of 122,516 examples for $\mathbf{T}$ in order to make $\mathbf{T}$ satisfy the first condition as much as possible. As we mentioned in section II-D, we drove the car as close as possible to the lane center. Figure 5-(a) shows the normalized histograms of the steering commands in $\mathbf{S}$ and $\mathbf{T}$. It is seen in the figure that $\mathbf{S}$ has more steering commands of large absolute values than $\mathbf{T}$. The source network trained with $\mathbf{S}$ drives roads in the source domain very well. 

To evaluate the driving performance of a trained CNN, we select four driving courses in GTAV. Figure 6-(a) and (b), respectively, show the example road images and the bird's-eye views of the four courses. Each course has a length of 2.5km and has different road characteristics. For examples, $\textit{course 1}$ is mostly straight and consists of asphalt roads, cement roads, and tunnels. In contrast, $\textit{course 4}$ consists of asphalt roads and is mostly curvy. We let a CNN drive each course three times. Therefore, the CNN drove a total of 30km. The driving information (steering command, yaw rate, and two dimensional position of the car) was also recorded at 10Hz while driving. Note that the speed of the car is controlled by a simple algorithm that tries to keep a car running at a constant speed, 30km/h. Our driving and data collection systems for TORCS and GTAV are based on \cite{Chen} and \cite{vpilot}, respectively.

\subsection{Performance Comparison}
We compare the proposed network with three CNNs trained with different methods:\\
$~~$ $\bullet~\textit{baseline}$~(BL) : the CNN ($\theta_{conv}$, $\theta_{f1}$) trained with $\mathbf{T}$ by using the following MSE loss
\begin{equation}
\mathcal{L}(\mathbf{x}, \mathbf{y}; \theta_{conv}, \theta_{f1}) = ||\mathbf{y} - f(\mathbf{x}; \theta_{conv}, \theta_{f1})||^{2}
\end{equation}
$~~$ $\bullet~\textit{source network}$~(SN) : the CNN ($\theta_{conv}$, $\theta_{f1}$) trained with $\mathbf{S}$ by using the loss in (4)\\
$~~$ $\bullet~\textit{domain transfer network}$~(DTN): the CNN ($\theta_{conv}$, $\theta_{f1}$) trained with the union of $\mathbf{S}$ and $\mathbf{T}$  by using both the loss in (4) and the domain confusion loss of \cite{Tzeng}\\
Note that all the parameters were initialized by Xaiver method \cite{Xavier} and optimized via Adam \cite{Adam} with the initial learning rate $\gamma=0.0001$. Each network was trained for 100 cycles through the number of the examples in the corresponding dataset, and the updated parameters with the best prediction performance over the validation dataset were chosen. Note that we let the size of the validation dataset be 5 percent of the training dataset.
\begin{table}[t]
\centering
\caption{Average VYR Performance}
\label{table2}
  \begin{tabular}{| c c c c c c | }
    \hline
    Course Num.    & BL       & SN     & DT      & Prop.1      & Prop.2                \\ \hline \hline
    1   		   & -        & 3.14   & 3.23    & 0.66        & 0.88        \\ \hline
    2    		   & -        & 1.45   & 7.54    & 0.70        & 0.71        \\ \hline
    3      		   & -        & 1.5    & -       & 0.71        & 0.48        \\ \hline
	4       	   & -        & 1.77   & -       & -           & 2.34        \\ \hline
    average        & -        & 1.96   & -       & -           & 1.1        \\ \hline
  \end{tabular}
\end{table}
\begin{table}[t]
\centering
\caption{Average D2GT Performance (meter)}
\label{table3}
  \begin{tabular}{| c c c c c c | }
    \hline
    Course Num.    & BL       & SN     & DT      & Prop.1       & Prop.2                \\ \hline \hline
    1   		   & -        & 0.78   & 1.11    & 0.99        & 0.75        \\ \hline
    2    		   & -        & 0.97   & 4.21    & 0.85        & 0.92        \\ \hline
    3      		   & -        & 0.89   & -       & 1.09        & 1.03        \\ \hline
	4       	   & -        & 0.75   & -       & -           & 0.67        \\ \hline
    average        & -        & 0.85   & -       & -        & 0.84        \\ \hline
  \end{tabular}
\end{table}

For objective comparisons, we propose two measures: 1) variance of yaw rates (VYR), 2) distance to ground-truth trajectory (D2GT). The ground-truth trajectory is the set of two dimensional positions of the car, which we obtained while driving the car as close as possible to the lane center. VYR at time index $k$ is calculated as
\begin{equation}
VYR(k) = \frac{1}{L}\sum_{t=-l}^{l} yr(k+t)^{2} - \left( \frac{1}{L}\sum_{t=-l}^{l} yr(k+t) \right) ^{2},
\end{equation}
where $yr(k)$ denotes the yaw rate of the car recorded at time index $k$ and $L=2l+1$. We set $l=5$ in this paper so that $VYR(k)$ is the variance of the yaw rates during 1 second. Let $(x_{k}, y_{k})$ denote the two dimensional position of the car at time index $k$. Also, let $(x_{g1}, y_{g1})$ and $(x_{g2}, y_{g2})$ denote the positions in the ground-truth trajectory, which are the first and the second closest positions to $(x_{k}, y_{k})$. Then, D2GT at time index $k$ is calculated by
\begin{equation}
D2GT(k) = |a_{g}x_{k} + b_{g}y_{k} + c_{g}| \times ( a_{g}^{2} + b_{g}^{2} )^{\frac{-1}{2}},
\end{equation}
where $a_{g}$, $ b_{g}$, and $c_{g}$ are the coefficients of the linear function, which passes $(x_{g1}, y_{g1})$ and $(x_{g2}, y_{g2})$. Figure 5-(b) shows an example of calculating D2GT values at time index $k$. Small VYR indicates that the car was controlled stably while small D2GT indicates that the car was driven safely along the lane center. 

Table 2 and 3 show the average of VYR and D2GT values, respectively. The values in the tables were obtained by first calculating the two measures at each time index and then averaging over the all recording times. Prop.1 and Prop.2 in the tables denote that the car was controlled by the proposed network according to $\hat{\mathbf{y}}^{t}=f(\mathbf{x}; \hat{\theta}_{conv}, \hat{\theta}_{f3})$ and $\hat{\mathbf{y}}^{share}$ in (3), respectively. Blank spaces in the tables mean that the car left the lane for a long time so that we cannot calculate the measures for the whole section of the course.

For subjective comparisons, we show the recorded trajectories of the car controlled by each CNN in Figure 7 and 8. The blue line denotes the ground-truth trajectory while the orange, green, and red lines denote the trajectories obtained from BL, SN, and Proposed 2, respectively. More subjective comparison results can be found at \url{http://ddokkddokk.tistory.com/64}.

We can see in the table that BL completed none of the four courses while Prop.1 completed course 1$\sim$3. While driving the three courses, the car controlled by Prop.1 left the lane center occasionally, but returned to the center. This implies that the loss function in (2) worked well. On course 4, however, the car failed to return to the lane center. This is because most of the roads in the target domain are straight or gently curvy so the cases of deviating from the lane center take up a small part of $\mathbf{T}$. On the other hand, SN completed all the courses. It showed good driving performances on $\textit{course 4}$ as seen in Figure 8 since the roads in $\textit{course 4}$ look similar to the roads in the source domain. However, SN showed unstable steering wheel controls on the roads of $\textit{course 1}$ $\sim$ $\textit{3}$, which are unfamiliar with SN (e.g. cement roads, tunnels, forks in roads) as seen in Figure 7. Finally, Prop.2 shows stable steering wheel controls on course 1$\sim$3 while successfully completed course 4. As a result, it shows the best performance on the four courses on average. Prop.2 can be regarded as the reasonable compromise between SN and Prop.1.

\section{Conclusion}
This paper proposed a method of transferring the driving experience a CNN has learned on the source domain to a target network for the task in the target domain. The experience transfer was accomplished through LDL predicted by the source network, and is based on our observation that the source network well recognizes common information (such as lane lines) between the source and the target images. The proposed network successfully drove on all the courses in the target domain while the others failed to complete the courses or showed unstable steering wheel controls.

Our future works may include driving a real car on highways by using the proposed method.

\addtolength{\textheight}{-12cm}   






\begin{thebibliography}{99}

\bibitem{Pomerleau} D. A. Pomerleau, ÒAlvinn: An autonomous land vehicle in a neural network,Ó Technical report, DTIC Document, 1989.
\bibitem{Huval} B. Huval, T Wang, S. Tandon, J. Kiske, W. Song, J. Pazhayampallil, M. Andriluka, P. Rajpurkar, T. Migimatsu, R. C.-Y., F. Mujica, A. Coates, and A. Y. Ng, ÒAn empirical evaluation of deep learning on highway driving,Ó arXiv preprint arXiv:1504.01716, 2015.
\bibitem{Chen} C. Chen, A. Seff, A. Kornhauser, and J. Xiao, ÒDeepdriving: Learning affordance for direct perception in autonomous driving,Ó in Proceedings of the IEEE International Conference on Computer Vision, pp 2722-2730, 2015.
\bibitem{Bojarski} M. Bojarski, D. D. Testa, D. Dworakowski, B. Firner, B. Flepp, P. Goyal, L. D. Jackel, M. Monfort, U. Muller, J. Zhang, X. Zhang, J. Zhao, and K. Zieba, ÒEnd-to-end learning for self-driving cars,Ó arXiv preprint arXiv:1604.07316, 2016.
\bibitem{Xu} H. Xu, Y. Gao, F. Yu, and T. Darrell, ÒEnd-to-end learning of driving models from large-scale video datasets,Ó in Proceedings of the IEEE International Conference on Computer Vision and Pattern Recognition, pp 2774-2182, 2017.
\bibitem{Lee} S. Lee, J. Kim, J. S. Yoon, S. Shin, O. Bailo, N. Kim, T.-H. Lee, H. S. Hong, S.-H. Han, and I. S. Kweon, ÒVPGNet: Vanishing point guided network for lane and road marking detection and recognition,Ó in Proceedings of the IEEE International Conference on Computer Vision, pp 1947-1955, 2017.
\bibitem{Yu} H. Yu, S. Yang, W. Gu, and S Zhang, ÒBaidu driving dataset and end-to-end reactive control model,Ó in Proceedings of the IEEE Intelligent Vehicles Symposium, pp 341-346, 2017.
\bibitem{Daume1} H. Daume III and D Marcu, ÒDomain adaptation for statistical classifiers,Ó Journal of Artificial Intelligence Research, vol. 26, pp. 101-126, 2006.
\bibitem{Daume2}  H. Daume III, ÒFrustratingly easy domain adaptation,Ó arXiv preprint arXiv:0907.1815, 2009.
\bibitem{Aytar} Y. Aytar and A. Zisserman, ÒTabula Rasa: Model transfer for object category detection,Ó in Proceedings of the IEEE International Conference on Computer Vision, pp 2252-2259, 2011.
\bibitem{Fernando} B. Fernando, A. Habrard, M. Sebban, and T. Tuytelaars, ÒUnsupervised Visual Domain Adaptation Using Subspace 
Alignment,Ó in Proceedings of the IEEE International Conference on Computer Vision, pp 2960-2967, 2013.
\bibitem{Chopra} S. Chopra, S. Balakrishnan, R. Gopalan, ÒDLID: Deep learning for domain adaptation by interpolating between domains,Ó in Proceedings of ICML workshop on Challenges in Representation Learning, 2013.
\bibitem{Hoffman} J. Hoffman, S. Guadarrama, E. Tzeng,  R. Hu, and J. Donahue, ÒLSDA: large scale detection through adaptation,Ó in Proceedings of Neural Information Processing Systems, 2014.
\bibitem{Tzeng} E. Tzeng, J. Hoffman, T. Darrell and K. Saenko, ÒSimultaneous deep transfer across domains and tasks,Ó in Proceedings of the IEEE International Conference on Computer Vision, pp 4068-4076, 2015.
\bibitem{Fernando2} B. Fernando, and T. Tommasi, and T. Tuytelaars, ÒJoint cross-domain classification and subspace learning for unsupervised adaptation,Ó Pattern Recognition Letters, vol. 65, pp. 60-66, 2015.
\bibitem{Bousmalis} K. Bousmalis, G. Trigeorgis, N. Silberman, D. Krishnan, and D. Erhan, ÒDomain separation networks,Ó in Proceedings of Neural Information Processing Systems, 2016.
\bibitem{Pan} X. Pan, Y. You, Z. Wang, and C. Lu, ÒVirtual to real reinforcement learning for autonomous driving,Ó arXiv preprint arXiv:1704.03952, 2017.
\bibitem{vpilot} Òhttps://github.com/aitorzip/VPilotÓ
\bibitem{Wymann} Òhttp://www.torcs.orgÓ
\bibitem{Zhou} B. Zhou, A. Khosla, A. Lapedriza, A. Olivia, and A. Torralba, ÒLearning deep features for discriminative localization,Ó in Proceedings of the IEEE International Conference on Computer Vision and Pattern Recognition, pp 2921-2929, 2016.
\bibitem{Xavier} X. Glorot and Y. Bengio, ÒUnderstanding the difficulty of training deep feedforward neural networks,Ó in Proceedings of International Conference on Artificial Intelligence and Statistics, 2010.
\bibitem{Adam} D. Kingma and J. Ba, ÒAdam: a method for stochastic optimization,Ó arXiv preprint arXiv:1412.6980, 2014.

\end{thebibliography}
\end{document}